\documentclass[pdflatex,sn-mathphys]{sn-jnl}
\usepackage{arabtex}
\usepackage{utf8}
\usepackage{lineno}

\algnewcommand\algorithmicforeach{\textbf{for each}}
\algdef{S}[FOR]{ForEach}[1]{\algorithmicforeach\ #1\ \algorithmicdo}

\jyear{2022}%

\theoremstyle{thmstyleone}%
\theoremstyle{thmstyletwo}%

\theoremstyle{thmstylethree}%

\begin{document}

\title{A Human Word Association based model for topic detection in social networks}


\author[1]{\fnm{Mehrdad} \sur{Ranjbar-Khadivi}}\email{mehrdad.khadivi@iaushab.ac.ir}

\author*[1]{\fnm{Shahin} \sur{Akbarpour}}\email{akbarpour@iaushab.ac.ir}

\author*[2]{\fnm{Mohammad-Reza} \sur{Feizi-Derakhshi}}\email{mfeizi@tabrizu.ac.ir}

\author[1]{\fnm{Babak} \sur{Anari}}\email{anari@iaushab.ac.ir}

\affil[1]{\orgdiv{Department of Computer Engineering}, \orgname{Shabestar Branch, Islamic Azad University}, \orgaddress{\city{Shabestar}, \state{East Azerbaijan}, \country{Iran}}}

\affil[2]{\orgdiv{Computerized Intelligence Systems Laboratory, Department of Computer Engineering}, \orgname{University of Tabriz}, \orgaddress{\street{29 Bahman Blvd.}, \city{Tabriz}, \state{East Azerbaijan}, \country{Iran}}}

\abstract{}
With the widespread use of social networks, detecting the topics discussed on these platforms has become a significant challenge. Current approaches primarily rely on frequent pattern mining or semantic relations, often neglecting the structure of the language. Language structural methods aim to discover the relationships between words and how humans understand them. Therefore, this paper introduces a topic detection framework for social networks based on the concept of imitating the mental ability of word association. This framework employs the Human Word Association method and includes a specially designed extraction algorithm. The performance of this method is evaluated using the FA-CUP dataset, a benchmark in the field of topic detection. The results indicate that the proposed method significantly improves topic detection compared to other methods, as evidenced by Topic-recall and the keyword F1 measure. Additionally, to assess the applicability and generalizability of the proposed method, a dataset of Telegram posts in the Persian language is used. The results demonstrate that this method outperforms other topic detection methods.


\maketitle


\section{Introduction}\label{sec:introduction}
As the volume of unstructured data generated by IoT devices, humans, and social media platforms grows, data science has emerged as a powerful analytical discipline for extracting valuable insights from this data \cite{Tien2017, Zehtab2023}. Social networks can be considered a type of complex network \cite{REIHANIAN2018370}. Analyzing social media content can help understand public events and people's opinions, concerns, and expectations \cite{COVIDTrendingTopics}. Identifying and investigating trending topics is an effective way to understand the nature of emerging issues genuinely.

Information is data that has been interpreted by humans using specific means \cite{Shi2011}. Valuable information can be obtained by examining and analyzing the rich and continuous flow of content within these media. Social media differ from other forms of media; for instance, their content is usually short and unstructured.

Among the various methods proposed for identifying topics in social networks, feature-based topic detection methods consider a topic to be a group of words that occur simultaneously and consecutively. These methods employ different approaches to identifying these keywords. For example, frequent pattern mining (FPM) has been utilized to identify topics on Twitter. The aim of FPM in topic detection is to find patterns of words that repeatedly occur in a set of documents \cite{Aiello2013}. However, existing methods are limited in some aspects. FPM methods cannot track the order of items as they are applied to transactions and item sets. Newer methods consider semantic relations, but they ignore the structure of language and linguistics.

Topic detection is one of the natural language processing (NLP) methods, so it is important to pay more attention to the structure of the language. The structure of the language refers to how humans understand it. In this paper, linguistic methods have been employed to extract these frequent patterns. A promising way to enhance these methods is to discover the association between keywords. To understand the human ability to identify word associations, preliminary work in the field of human word association has been conducted by Klahold et al. \cite{Uhr2013, Klahold2014}. Based on the results of case studies, this paper utilizes the concept of "imitating the mental ability of word association."

Klahold et al. have performed topic extraction on long and structured texts, while this article uses this method on microblogs. For this purpose, a model based on human understanding of word association, named Human Word Association (HWA), has been proposed. The functionality of this method and an explanation of how it is used to uptake human word understanding are given in Section \ref{sec:HWA}.

Another fundamental problem is the generation of repetitive topics with the same meaning. This problem can be addressed by using the structure of the language. For this purpose, this paper presents a special extraction algorithm that utilizes embedding and clustering methods. The main contributions of this work include:
\begin{itemize}
\item Using the HWA method for topic detection in social networks
\item Proposing a new pattern extraction algorithm using AGF to address the issue of producing repetitive synonym patterns
\item Proposing a new keyword rating formula based on word score and utility
\item Applying this framework to the Persian language
\end{itemize}

The structure of this paper is as follows: Section \ref{sec:related_work} reviews related works in this field. Section \ref{sec:HWA} presents the concept of human understanding of word association. The details of the proposed model are given in Section \ref{sec:proposed_method}. Section \ref{sec:exp-results} provides an overview of the dataset, evaluation criteria, and experimental results. Finally, the paper is concluded in Section \ref{sec:conclusion}.

\section{Related Works} \label{sec:related_work}
The original idea of topic detection and tracking was initiated in 1996 by the Defense Advanced Research Projects Agency (DARPA) for the Broadcast News Understanding program \cite{EnhancedHbGraph}. This idea has been developed over the years using different techniques. Different approaches for Topic detection in text data streams, such as online social networks, differ in how they group and analyze the data based on the semantics or the distributions of the text. This practice is divided into two major categories: document pivot and feature pivot \cite{Indra2018, HUPM}. With the advent of social networks, monitoring and analyzing this rich and continuous flow of user-generated content can yield valuable and unprecedented information unavailable in traditional media. This field is well-studied, and good review articles have been written \cite{Gaglio2015, Asgari-Chenaghlu2021, Liu2012, Saeed2019}.
\subsection{document-pivot approaches}\label{subsec:doc-p}
The document-pivot approach is a topic detection technique that groups microblog posts (like tweets) with similar meanings and assigns them a topic. These methods usually use approaches that cluster documents based on similarity. For this purpose, a semantic distance between documents is needed \cite{Cordeiro2016}. Document-pivot methods usually recognize bursty \footnote{has a higher occurrence frequency than others} states based on clustered topics \cite{Zong2021}. This technique was developed based on First Story Detection (FSD) research and Local Sensitivity Hashing (LSH). Petrovic et al. in \cite{FSD} used TF-IDF and co-occurrence of the document words to find the similarity. They used Umass \cite{UMass} as the basic clustering. 
Document-pivot methods can better capture the semantics of the documents, but they may suffer from high computational complexity and noise. For example, they are only based on the co-occurrence of words and these methods fail if frequent words have different meanings.  

\subsection{feature-pivot approaches}\label{subsec:feat-p}
The feature-pivot approach performs document clustering based on features. study the distributions of words and discover events by grouping words \cite{Cordeiro2016}. For example, a feature-pivot method might identify keywords with a sudden increase in frequency or co-occurrence and use them to represent a topic. Feature-pivot methods usually recognize bursty states based on feature discovery \cite{Zong2021}. This approach is divided into feature-based and probabilistic topic models \cite{COVIDTrendingTopics, HUPC, HUPM}. In feature-based, a topic is a group of words that are based on the threshold values approach. In contrast, the probabilistic topic model uses a feature called probability distribution. Also, with the emergence of newer methods based on machine learning and combining them with topic detection methods, a third group has been added to this division. We named this group novel or hybrid methods.

\subsubsection{feature-based}
Most scientific achievements in topic modeling are based on Dirichlet's Latent Allocation (LDA) \cite{Saeed2020}. LDA is a probabilistic model built on BoW (Bag of Words) and widely used for topic modeling. The amount of word repetition is extracted from the documents to calculate the probability distribution of the words likely found in the topics \cite{Blei2003}. O’Connor et. al. in  \cite{tweetmotif} presented the Graph-Based Feature-Pivot (Gfeat-p) topic identification technique. As a result, every text is converted into a graph, and the Structural Algorithm for Networks (SCAN) is used to compute the clusters. This approach looks at the relationships between the words and examines related graphs to find topics.

A problem with feature-pivot methods is that to group a set of terms, they only consider feature pairwise similarities. If closely interconnected topics share a relatively large number of terms, this procedure will most likely produce generic or merged topics \cite{Aiello2013}. An option to deal with this issue is to find patterns or associations that occur frequently. The most widely used technique is to use frequent pattern mining. The problem of frequent pattern mining has been well studied due to its various applications in many fields, such as clustering and categorization. Frequent pattern mining is a common method for topic detection problems on the Twitter data stream, and various methods have been proposed. \cite{Aiello2013} proposed the BNgram approach, which uses n-grams instead of unigrams for topic detection. The idea is that the topics may have recurring structures (like retweets). The computation is done using DF-IDF, a useful score for identifying recurring and related patterns. Furthermore, the application of Name Entity Recognition (NER) illustrates the significance of proper nouns in event detection. In \cite{SFPM}, the paper's authors designed a soft recurrent pattern mining (SFPM) approach to overcome the problem of topic detection. This research aims to find the co-occurrence of words with a value greater than two, so the probability of each word is calculated separately. After finding the most frequent K (top-K), a co-occurrence vector is formed to add the co-occurring words to the top-K word vectors \cite{SFPM}. As an improvement for this method, in \cite{Gaglio2015}, a system is designed to overcome the limitation of SFPM in dealing with dynamic and real-time scenarios \cite{Asgari-Chenaghlu2021}. In \cite{Exemplar-Based} an exemplar-based approach is proposed. The goal is to find tweets that provide valuable context for a topic. This method's concept is that any occurrence can be summed up in a tweet. Select tweets that have the greatest overlap with related tweets of a topic and the least overlap with tweets about other topics are designated as representations. The approach known as Separable Non-negative Matrix Factorization (SNMF), first presented by \cite{SNMF}, breaks apart the matrices to produce the terms and themes matrix, after which the events are identified by that matrix. This method uses KL recovery as part of the algorithm, in addition to the original recovery, which makes use of algebraic manipulation. The document matrices become factorized matrices, which are subsequently grouped, in the method described by \cite{SVDKmeans}, which combines Singular Value Decomposition (SVD) with K-means. After extracting each cluster center, associated keywords that describe the events are extracted.

\subsubsection{probabilistic topic models}

The event detection system "Twevent" \cite{twevent} first extracts contiguous and non-overlapping word segments (single words or phrases) from each tweet. The obtained statistical information is used to detect unnecessary parts of the word. Then the top-k segments of the burst event are calculated in a fixed time window from the frequency of the burst segments with the user frequency of the burst segments. Finally, a variant of the Jarvis-Patrick clustering algorithm is used for grouping. Events are then filtered based on their newsworthiness score. The events this algorithm detects heavily depend on the Microsoft Web service N-gram and Wikipedia. This dependency can result in a series of events affected by the service. Additionally, events that have yet to be reported on Wikipedia may not be identified.

\cite{Liu2012} introduces pattern mining with High Utility. This approach has been used in topic detection approaches like \cite{HUPC, HUPM}. In \cite{HUPC}, a method called HUPC, after determining the usage rate of each pattern and extracting patterns with high utility, top-K very similar patterns are selected using the modularity-based clustering method and the KNN classification method. Also, in \cite{HUPM}, HUPM was used to find a group of frequently used words, and then a data structure called TP-tree was used to extract the pattern of the main topic.

Saeed et al. in \cite{EnhancedHbGraph, Saeed2018, Saeed2019, Saeed2020} proposed a dynamic heartbeat graph (DHG) technique to detect topics. This algorithm creates a subgraph for each sentence. These subgraphs are then added to the main graph, and an operation is applied to the entire graph based on the co-occurrence of words (edges between nodes) \cite{Asgari-Chenaghlu2021}.

\subsubsection{novel or hybrid methods}

Most of the mentioned methods are based on the frequency of patterns or pay attention to the co-occurrence of words. however, The semantic relationship between words must be addressed. Asgari et al. in \cite{TopicBert} use the combination of transformers with the incremental community detection algorithm to detect topics. On the one hand, transformers provide the semantic relationship between words in different contexts. On the other hand, the graph mining technique improves the accuracy of the resulting topics with the help of simple structural rules. This approach uses a novel embedding technique named BERT for creating graphs. As a hybrid method, a Semantic Modular Model (SMM) with five distinct modules—Distributional Denoising Autoencoder, Incremental Clustering, Semantic Denoising, Defragmentation, and Ranking and Processing—is proposed by Hadizadeh Moghaddam and Momtazi in \cite{SMM}. The suggested model seeks to: (1) group different documents together and eliminate those that might not be useful in identifying events; and (2) pinpoint more significant and illustrative keywords. 

It should be noted that the topic detection problem is an NLP problem, but the linguistic methods have yet to be investigated in the previous methods. Therefore, language structure and text processing methods will improve the results. One of these methods is imitating the human ability of word understanding, called the Human Word Association (HWA). Preliminary studies in this field have been carried out by Klahold et al. \cite{Aiello2013, Klahold2014}.

\section{Human Word Association} \label{sec:HWA}
Traditional methods are limited to calculating the frequency of patterns and co-occurrence extraction. Although newer methods pay attention to semantic relations, they need to consider the structure of the language. A study in psychology and linguistics shows that a significant part of human understanding of the relationship between words is asymmetric, meaning that the strength of the relationship between two words is different based on their order.

Nelson et al. \cite{Steyvers2004, Nelson2000, Nelson2004} analyzed the human understanding of word relations. They conducted a test with six thousand participants. In this test, called the Free Association Test (FAT), designed by Russell \cite{Russell1970}, and the Free Association Norm (FAN) presented by Nelson et al. \cite{Nelson2000}, stimulus words are presented to the participants. They are asked to write the word that comes to their mind related to the given stimulus word. Nelson claimed that the pair (good, bad) is symmetrical because about 75\% of the participants answered "good" as a response to the word "bad," and about 76\% of them answered the word "bad" in response to the word "good". While 69\% of people provided the word "bird" as an answer to the word "canary," only 6\% of the participants gave the word "canary" as an answer to the word "bird". So, the pair (canary, bird) is asymmetric \cite{Klahold2014}. Therefore, for successful simulation, the proposed method should cover the asymmetric character of HWA. In this regard, Uher et al. \cite{Uhr2013, Klahold2014} presented a concept to imitate the ability of the human mind to understand the relationship between words, called CIMAWA. This concept, on the one hand, and the power of the word itself, on the other hand, has led to the development of a formulation called the Association Gravity Force (AGF). This concept shows the power of a word concerning another word  \cite{Uhr2013, Klahold2014}.
This paper uses this concept, called HWA, to extract topics that better describe data flow in social networks. To make this concept practical in this framework, we conducted several experiments and in these experiments, we examined the co-occurrence rate of two words, the semantic distance in different language models, and the AGF of the words. For example, suppose the words “final” and “time” are from the last batch of the FA-Cup dataset. In that case, their co-occurrence rate equals 72, which ranks 535 (we calculated their co-occurrence rate and sorted them in descending order). The similarity based on the W2V model is equal to 0.49. However, the same amount based on AGF is equal to 5.55. It should also be noted that if we change the order of the words, the amount of co-occurrence and semantic similarity will not make a difference, while the amount of AGF will be different. The placement of words creates meaningful topics. This arrangement does not affect the FA-Cup dataset results, but it is important in the Telegram dataset due to the nature of the topics. In the following, the mechanism of the proposed method is given.

\section{HWA based Proposed Method} \label{sec:proposed_method}
\begin{figure}
    \centering
    \includegraphics[width=1\textwidth]{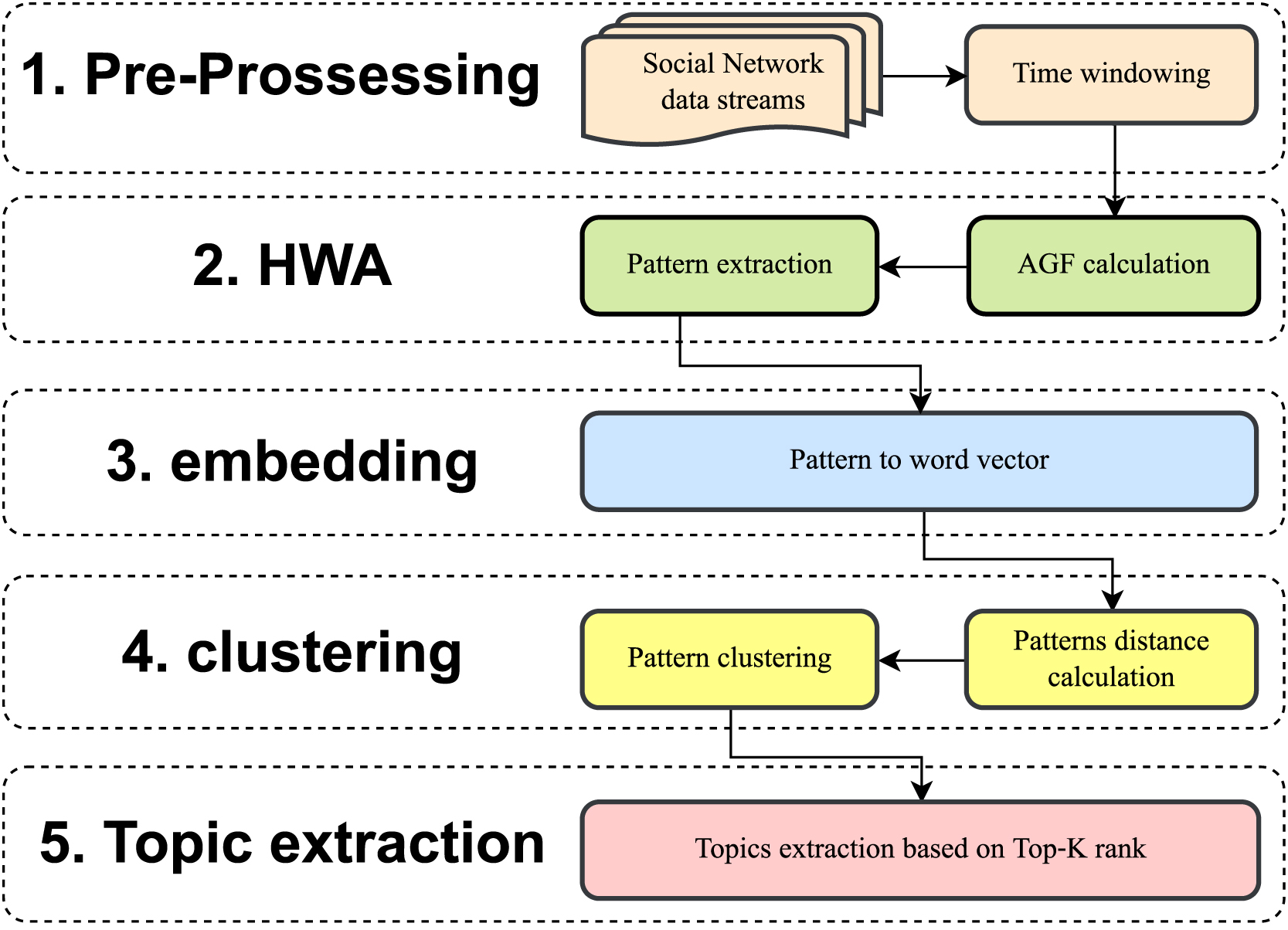}
    \caption{The flowchart of the proposed framework. This framework has 5 steps. In the first step, after receiving the data stream, pre-processing operations such as windowing and tokenizer are performed. HWA values are calculated in the second step, and patterns are extracted using the presented algorithm. In the next step, the word vector of each pattern is extracted so that in step 4, each pattern's similarity(distance) is calculated and the patterns can be clustered. Finally, in the last step, the topics are selected and extracted based on the top-k ranking}
    \label{fig:flowchart}
\end{figure}

The proposed algorithm uses all three natural language processing methods, including the co-occurrence of keywords, the semantic relationship between words, and the human understanding of word association, emphasizing linguistic structures. Figure \ref{fig:flowchart} represents The flowchart of the proposed algorithm. The data from social networks to topic extraction goes through five steps:
\begin{itemize}
  \item Receiving the social network data stream, windowing, and preprocessing
  \item Keyword ranking, co-occurrence calculation, AGF finding, and pattern extraction
  \item words Embedding calculation and pattern vector extraction
  \item pattern distance calculation and pattern clustering
  \item cluster ranking and topic extraction
\end{itemize}

Social network data flow can be considered as a sequence of posts that are received in chronological order. All posts received recently and within a fixed time window (12 hours) can be represented by a batch of posts ${Batch}^L=\{{Post}_p^L\}$, where ${Post}_p^L$ represents the post $p$ in the $L$th window. It should be noted that in this paper, each post means the text of that post after pre-processing and removing stop words. If ${Word}^L=\{{word}_1^L,{word}_2^L,...,{word}_w^L\}$ is the set of words of the window $L$, so ${Post}_p^L \subset {Word}^L$ will be. In other words, ${Post}_p^L$ is a set of words; each word is a member of the set of words of the same window. The final goal of this paper is to extract Topics from the window $L$ displayed as ${Topic}^L$.

Social network posts are highly likely to contain noise, such as misspelled words, emojis, mentions, hashtags, or URLs. This noise can cause many problems. Therefore, it is necessary to pre-process these posts and reduce the volume of these noises. The first step is to transform documents into a string of characters with various formats \cite{vasfisisi2013text}. For this purpose, a tokenizer written exclusively for processing social network posts has been used. This tokenizer separates tokens based on types of separators. It also recognizes the type of token. Many, but not all, types of tokens are emojis, URLs, hashtags, mentions, numbers, and simple or compound words. After tokenization, all tokens except numbers and words are removed from the posts. stop words are also left out.

After the pre-processing operation, keyword ranking calculation initiates the proposed method's steps. All words in the window will be ranked using the following steps.

\subsection{keyword rating}\label{subsec:keyword_rating}
A method for recognizing and ranking keywords is needed to identify frequent patterns to find the most important words in the time window. There are two approaches for this task, the first method is based on the tf-idf of each word, and the second method is based on increasing or decreasing the occurrence of that word in two consecutive windows. The first method underscores frequent words that occur in most posts, while some of them are very important for detecting topics. The second method is proposed to solve this problem. By calculating the occurrence trend of words (increase or decrease in occurrence) between two consecutive windows, this method selects the words whose trend is increasing as keywords. However, the trend of a word may be decreasing or constant while still being recognized as a keyword. To overcome these problems We propose a combined method based on both of these methods.

The rank of each word like $w$ in window $L$ is displayed by ${Kr}^L(w)$ and is calculated as equation \ref{eq:kr}.

\begin{equation}
        \label{eq:kr}
        {Kr}^L(w) =  \frac{{score}^L(w) + {utility}^L(w)}{2}
    \end{equation}

where $score^L(w)$ is the score of word $w$ in the $L^th$ window, which is calculated as equation \ref{eq:score}.

\begin{equation}
        \label{eq:score}
        {score}^L(w) = {TF}^L(w) \times \log{\frac{\|{Batch}^L\|}{{DF}^L(w)}}
    \end{equation}

where ${DF}^L(w)$ is the number of posts in the window $L$ that the word $w$ is appeared. Also, ${TF}^L(w)$ is the frequency of the word $w$ in window $L$. This is calculated using equation \ref{eq:TF}.

\begin{equation}
    \label{eq:TF}
    {TF}^L(w) = \sum_{p=1}^{\|{Batch}^L\|}  {TF}_p^L(w)
\end{equation}

where ${TF}_p^L(w)$ is the frequency of the word $w$ in post $p$ from window $L$.

Since there is a limited number of characters for any post, the number of specific words associated with a topic can quickly increase. To highlight this feature, a concept adapted from \cite{HUPM} named ${utility}_L(w)$ is defined by equation \ref{eq:utilityLw}. It is the utility of the word $w$ in the current window compared to the previous window.

\begin{equation}
    \label{eq:utilityLw}
    {utility}^L(w) = \begin{cases} 
                        {diff}^L(w) \times \log{\frac{{TF}^L(w) + 1}{{TF}^{L-1}(w) + 1}}, & {diff}^L(w) > 0 \\
                        0, &  {diff}^L(w) \leq 0
                    \end{cases} 
\end{equation}
where
\begin{equation}
        \label{eq:diffLw}
        {diff}^L(w) = {TF}^L(w) - {TF}^{L-1}(w)
    \end{equation}
    
${TF}^L(w)$ is calculated by equation \ref{eq:TF} and in the same way ${TF}^{L-1}(w)$ is the frequency of word $w$ in window $L-1$. If ${diff}^L(w) > 0$, word repetition is increasing. If ${diff}^L(w) < 0$, word repetition is decreasing.

In this paper, among the words in ${Word}^L$, the words that receive a higher $Kr$ value are considered keywords. Deciding how many of these words to select requires a threshold. Choosing a fixed threshold does not seem correct, considering that the number of words in each window is variable. Therefore, this paper will select $h\%$ of words with high $Kr$ as keywords. Various experiments have been conducted to achieve an optimal threshold. Based on the results of the experiments, the value of $h=30$ is used. Therefore, the words among the top $30\%$ are selected to continue and go to the next stage. In section \ref{subsec:ParamSetup} we discuss and illustrate the experiments that how we find these values. 

\subsection{Co-occurrence}\label{subsec:Cooc}
Co-occurrence is a standard method to estimate the correlation of words in a corpus. Co-occurrence refers to the probability that two words have appeared together multiple times in the same post. This paper calculates the co-occurrence of two keywords x and y if $x \neq y$ as equation \ref{eq:CooC}.

\begin{equation}
        \label{eq:CooC}
        {CooC}_p^L(x,y) = \begin{cases} \mbox{1} & \mbox{if }  x \in {Post}_p^L  \&  y \in {Post}_p^L \\
                            \mbox{0} & \mbox{otherwise} \end{cases}                                    x \neq y \bigwedge x, y \in {Word}^L
    \end{equation}

Consider that x and y are two different words from the set of words of the window $L$. If both words appear in a post from the same window, the co-occurrence value of these two words in that post is assumed to be one; otherwise, its value will be zero. The total co-occurrence of two words in the entire window $L$ is calculated as equation \ref{eq:CooC_tot}.

\begin{equation}
        \label{eq:CooC_tot}
        {CooC}^L(x,y) = \sum_{p=1}^{\|{Batch}^L\|}  {CooC}_p^L(x,y)
    \end{equation}

\subsection{CIMAWA}\label{subsec:CIMAWA}

To achieve the power of HWA in the way humans perceive it and the symmetric and asymmetric combination of word association, a concept called CIMAWA \cite{Klahold2014} has been used. It is as equation \ref{eq:CIMAWA}.

\begin{equation}
    \label{eq:CIMAWA}
    {CIMAWA}^L(x,y) = \frac{{CooC}^L(x,y)}{{TF}^L(y)} + \delta \times \frac{{CooC}^L(x,y)}{{TF}^L(x)}
\end{equation}

$\delta$ is a damping factor; its value is assumed to be 0.5. CIMAWA is used to calculate AGF in the next step.

\subsection{Associative Gravity Force(AGF)}\label{subsec:AGF}
Keyword extraction methods identify the keywords of the texts and rank them according to their importance, but they are limited in terms of the thematic structure of the text. To obtain a function that provides not only the keywords but also a topic about the corpus using those keywords, a definition called Association Gravity Force (AGF) \cite{Klahold2014} is used. This concept is defined as equation \ref{eq:AGF}.

\begin{equation}
    \label{eq:AGF}
    {AGF}^L(x,y) = {CIMAWA}^L(x,y) \times \frac{{Kr}^L(x)}{{Kr}^L(y)}
\end{equation}
    
\subsection{Pattern extraction using AGF}\label{subsec:Pattern_extraction}

Now, using AGF, the set of words that have the highest association with the word $w$ in window $L$ is defined as follows:
    
\begin{equation}
    \label{eq:high_AGF}
    M^L(w) = \{ a \| a \in {Word}^L \bigwedge a = \arg\max_{x \in {Word}^L} (AGF^L(w,x)) \} 
\end{equation}

'a's are the words in ${Word}^L$ whose AGF with word $w$ is maximum compared to other words. The table \ref{tbl:ML} gives an example of how to obtain these words. Since several words in ${Word}^L$ may have the highest AGF value with word $w$, The $M^L(w)$ is defined as a set.

\begin{table}[]
    \centering
    \caption{ An example of obtaining $M^L(w)$:
    \newline Assume that ${Word}^L=\{w, w_0, w_1, w_2, w_3, w_4, w_5, w_6, w_7, w_8, w_9\}$ and the value of AGF of word w with each word is calculated like the table. Considering that the maximum value of AGF equals 8, $M^L(w) = \{w_1, w_9\}$. A word may have the maximum value of AGF with more than one word. Therefore, $M^L(w)$ is considered a set.}
    \label{tbl:ML}
    \begin{tabular}{|c|c|}
        \hline
        Word & AGF \\
        \hline
        $w_0$ & 2 \\
        \hline
         $w_1$ & 8 \\
        \hline
         $w_2$ & 1 \\
        \hline
         $w_3$ & 5 \\
        \hline
         $w_4$ & 4 \\
        \hline
         $w_5$ & 7 \\
        \hline
         $w_6$ & 5 \\
        \hline
         $w_7$ & 6 \\
        \hline
         $w_8$ & 3 \\
        \hline
         $w_9$ & 8 \\
        \hline
    \end{tabular}
\end{table}

To identify topics, using $AGF$ and $M^L$, patterns will first be extracted in the way given in Algorithm \ref{alg:Pattern_extraction}. These patterns will be calculated first for each word. Using pattern extraction, all the words that can produce a pattern based on the concept of HWA be placed next to each other. Each pattern is a combination of different words with high communication power as ${Pattern}_j^L=\{w_z\}$, where $w_z \in {Word}^L$. By repeating the above process for all the words in ${Word}^L$, the set of patterns will be obtained as ${Patterns}^L = \{{Pattern}_j^L\}$. Since repetitive patterns may be produced during the process, patterns will be merged using post-processing operations. For example, ${Pattern}^{37}$ is a set of patterns for window 37, and two patterns from this set are 
${Patterns}_1^{37}$ = \{fire, Plasko, building, burn\} (\{ \setcode{utf8} \<آتش>, \<پلاسکو>, \<ساختمان>, \<آتش سوزی> \})
 And
 ${Patterns}_2^{37}$ = \{Plasko, fire, building, burn, incident\}  (\{ \setcode{utf8} \<آتش>, \<پلاسکو>, \<ساختمان>, \<آتش سوزی>, \<حادثه>\})
 . The two patterns are merged because the first pattern is a subset of the second pattern.
In general, the extracted patterns show the span of the topics and may also suffer from the correlation problem. This problem may lead to a decrease in the accuracy of topic extraction. For a better understanding, from Window 16, all three patterns
${Patterns}_1^{16}$ = \{\} \{ \setcode{utf8} \<هاشمی>, \<مرگ>, \<آیت الله> \}
,
${Patterns}_2^{16}$ = \{ \setcode{utf8} \<رفسنجانی>, \<رحلت>, \<آیت الله> \}
And
${Patterns}_3^{16}$ = \{ \setcode{utf8} \<رفسنجانی>, \<فوت>, \<آیت الله>, \<هاشمی> \}
  are showing the event "Ayatollah Hashemi's death due to a heart problem".
All of the words \setcode{utf8} \<مرگ>,  \setcode{utf8} \<فوت> and  \setcode{utf8} \<رحلت> has the same meaning "death". So, they have to be grouped. The best solution to solve the problem is to cluster the patterns. A semantic relationship is used to achieve a criterion for clustering, explained in the next section.
    
\begin{algorithm}
\caption{Pattern extraction using AGF}
\label{alg:Pattern_extraction}
\begin{algorithmic}[1]
\State \textbf{Input:} ${Word}^L$, $M^L$
\State \textbf{Output:} patterns
\State ${patterns}^L \Leftarrow NULL$
\ForEach {w \textbf{in} ${Word}^L$}
    \State $Candidates \Leftarrow NULL$
    \State $Pattern   \Leftarrow NULL$
    \State Add  ${w}$ to ${Candidates}$
    \While{$Candidates is not empty$}
        \State c = Candidate[0] 
        \State delete c from Candidate 
        \If{$c$ not in $pattern$}
            \State Add  $c$ to $pattern$
            \State get $M^L(c)$ based on eq. \ref{eq:CooC_tot}
            \State Add $M^L(c)$ to Candidate
        \EndIf
        \State append $Pattern$ to ${Patterns}^L$
    \EndWhile
\EndFor
\ForEach {$P_i$,$P_j$ \textbf{in} ${Patterns}^L$}
    \If{${P_i} \subset {P_j}$}
        \State delete ${P_i}$ from ${Patterns}^L$
    \EndIf
\EndFor
\end{algorithmic}
\end{algorithm}
    
\subsection{Using Embedding to cluster patterns}\label{subsec:Embedding}
This section aims to find patterns with high semantic relationships, cluster them, and return the final topics as output. This will prevent the creation of repetitive patterns and the scattering of topics. Clustering is the best unsupervised method for community detection based on the semantic relationship of a pattern. The first step in pattern clustering is extracting their equivalent vector. One of these methods is the embedding extraction of each pattern. Word embedding methods can be categorized into three types: matrix-based methods (such as TF-IDF matrix, Latent Semantic Analysis (LSA), and GloVe), cluster-based methods (like Brown), and neural network-based methods (including Neural Network Language Model (NNLM), Log-Bilinear Language model (LBL), C\&W, skip-gram, continuous bag-of-words model (CBOW), and FastText) \cite{Shi2022}. In this paper the pre-trained fasttext\footnote{\url{https://fasttext.cc/}} model \cite{bojanowski2016enriching} is used. fastText is an efficient model for representing word vectors using NGrams. In this paper, the model trained for the Persian language using CBOW, in dimensions of 300, with NGram of length 5, has been used. For this purpose, their equivalent vector is extracted for the words in each pattern, and then the vector representing the pattern is calculated in the form of equation \ref{eq:Embpatternj}.

\begin{equation}
    \label{eq:Embpatternj}
    Emb({Pattern}_j^L) = \frac{1}{\|{Pattern}_j^L\|} \sum_{z=1}^{\|{Pattern}_j^L\|} Emb(w_z)
\end{equation}

where $Emb(w_z)$ is the equivalent vector of the word $w_z$. The distance between them will be calculated for all patterns by calculating the vector of each pattern. This distance will be used to cluster these patterns. The distance between two patterns is calculated as equation \ref{eq:dispatterns}.
    
\begin{equation}
    \label{eq:dispatterns}
    D_c ({Pattern}_j^L, {Pattern}_k^L) = 1 - S_c ({Pattern}_j^L,{Pattern}_k^L)
\end{equation}

$S_c$ shows the cosine similarity of two patterns and is calculated as equation \ref{eq:Sc}.

\begin{equation}
    \label{eq:Sc}
    S_c({Pattern}_j^L, {Pattern}_k^L) = \frac{Emb({Pattern}_j^L)\cdot Emb({Pattern}_k^L)}{\|Emb({Pattern}_j^L)\| \|Emb({Pattern}_k^L)\|}
\end{equation}

\subsection{patterns clustering and Topic extraction}\label{subsec:clustering}
When the patterns are extracted and their embedding is calculated, a clustering algorithm is used to cluster the obtained patterns based on the embedding obtained from the previous step and the cosine distance between them. Various algorithms can be used. The algorithm used in this paper is the Hierarchical Density-Based Spatial Clustering of Applications with Noise (HDBSCAN) algorithm. HDBSCAN is a clustering algorithm developed by Campello et al. \cite{HDBSCAN}. Due to the nature of social network data, it is impossible to use classical clustering algorithms. A classical algorithm such as K-means fails to group the data into useful clusters, even knowing the correct and expected number of clusters\cite{Shi2007}. On the other hand, HDBSCAN ensures optimal clustering. We have studied several clustering algorithms in our previous works \cite{khadivi2023GE}, and the results show that the HDBSCAN algorithm performs better than other methods.

\section{Experiments and Results}\label{sec:exp-results}

This section presents the experimental results obtained by the proposed framework. All the implementations were done in Anaconda Python 3.8 and ran on a PC with a 3.60GHz Core i7-7700 processor with 32 GB RAM and a Linux Mint operating system. The performance of the proposed work is compared with the results of the other models. The results announced by previous articles have been used to evaluate the FA-Cup dataset. While there are no previous results for the Telegram dataset, the methods have been implemented by authors and evaluated on this dataset using the approach described by these papers. The two datasets are described in section \ref{subsec:dataset} and the evaluation method is explained in section \ref{subsec:evaluation_metrics}.


\subsection{Dataset}\label{subsec:dataset}
Two datasets have been used to evaluate the proposed algorithm's performance. The first dataset, FA-Cup, is a benchmark dataset introduced by \cite{Aiello2013} in 2013 and collected from Twitter. The second data set is the Telegram dataset, which is used in this research to evaluate the effectiveness of the proposed method on the Persian language.
\subsubsection{FA-Cup dataset}\label{subsec:Dataset_facup}
FA Cup (Football Challenge Cup) is one of English football's oldest and most famous knockout competitions. The Fa Cup dataset, collected and published by Aiello et al. \cite{Aiello2013}  in 2013, contains data about the 2012 final match between Chelsea and Liverpool. This dataset consists of 360 one-minute time windows, 13 of which are ground-truth, including 13 topics, including the start, halftime, end of the match, and important game events. This dataset is one of the Benchmark datasets in the field of topic detection  \footnote{\url{http://www.socialsensor.eu/results/datasets/72-twitter-tdt-dataset}}. Many recent studies have used this dataset to evaluate their methods. The paper's authors did the windowing of this dataset(Aiello et al.).
According to Twitter's policies, publishing the content of tweets is not allowed, and the producers have only published the ID of the tweets. For this reason, the tweets have been restored by this paper's author. However, some tweets may not be available during collection for various reasons, such as deletion by the user. For this, the Twitter data stream API is used. Thus, the statistical information about the data set up to the moment of conducting the research is given in Table  \ref{tbl:stat datasets}.

\subsubsection{Telegram dataset}\label{subsec:Dataset_Telegram}

\begin{figure}
     \centering
     \includegraphics[width=\textwidth]{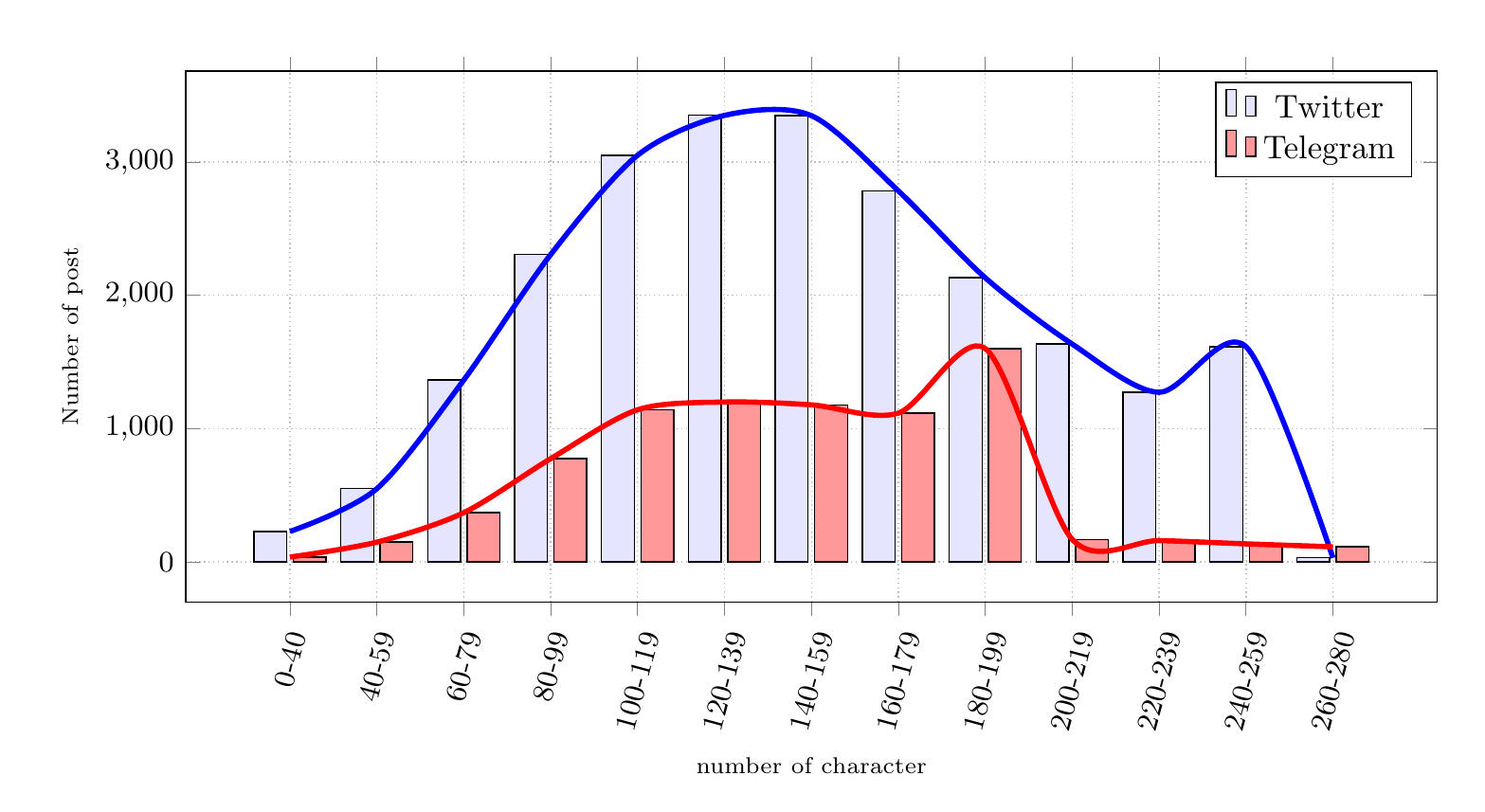}
     \caption{The number of characters of posts in Telegram and Twitter. Each microblog post usually contains less than 300 characters. This chart illustrates that both Twitter and Telegram can be considered a microblog.}
     \label{fig:Histo}
\end{figure}

The efficiency of a method is determined when that method can be used for different datasets. To review the generalizability of the method and its application on other social media platforms or other languages, we used the Telegram dataset. The Telegram dataset named $Sep\_General\_Tel01$ \cite{Sep-TD-Tel}, provided by the Computerized Intelligence Systems Laboratory\footnote{\url{cominsys.ir}} (ComInSyS). This dataset has been collected due to the lack of datasets in the Persian language and the high popularity of the Telegram social network in Iran. Telegram network is considered a microblog. Microblogs are short blog posts that are usually less than 300 words long and can contain images, GIFs, links, infographics, videos, and audio clips. Figure \ref{fig:Histo} shows the character length histogram of Telegram and Twitter posts. This histogram shows that although the maximum character length in Telegram is 4096 characters, most posts shared on Telegram are between 60 and 300 characters long. This feature is also available on other microblogging platforms such as Twitter. Thus, Telegram, like other social networks, will be a very good source for data analysis, especially in topic detection. The statistical information about this dataset is given in Table \ref{tbl:stat datasets}.

 In this work, the official API published by Telegram has been used. The implemented framework can retrieve text and media (images, documents, videos, etc.). To comply with the principle of privacy, only data related to public channels and groups have been used for collection. According to the studies, unlike existing datasets often collected from the Twitter social network and requiring keywords to retrieve information, the Telegram network does not require any keywords. This dataset was also collected without using any keywords. This data set contains more than ten thousand records of messages sent in public channels and groups in one month between 1395/10/12 (January 1, 2017) and 1395/11/12 (January 31, 2017). 
This dataset includes two super hot topics in the above time frame: "Death of Ayatollah Hashemi Rafsanjani" and "Plasco Building Fire". According to the nature of this research, only the text of the messages has been used. To process posts, this dataset is divided into 60 12-hour windows. To evaluate and compare the results of this paper, it is necessary to compare the extracted issues with Ground Truth (GT). For this purpose, among the existing sixty windows, nine windows that have the most news value and best match with two super topics have been selected for labeling. These windows are {14, 15, 16, 17, 18, 37, 38, 39, 40}. To perform this action, four experts were used for labeling, and the final label was the result of these people's opinions.

In this work, the official API published by Telegram has been used. The implemented framework can retrieve text and media. To comply with the principle of privacy, only data related to public channels and groups have been used for collection. According to the studies, unlike existing datasets often collected from the Twitter social network and requiring keywords to retrieve information, the Telegram network does not require any keywords. This dataset was also collected without using any keywords. This data set contains more than ten thousand records of messages sent in public channels and groups in one month between 1395/10/12 (January 1, 2017) and 1395/11/12 (January 31, 2017). This dataset includes two super hot topics in the above time frame: "Death of Ayatollah Hashemi Rafsanjani" and "Plasco Building Fire". 
According to the nature of this research, only the text of the messages has been used. To process posts, this dataset is divided into 60 12-hour windows. 

To evaluate and compare the results of this paper, it is necessary to compare the extracted issues with Ground Truth (GT). For this purpose, among the existing sixty windows, nine windows that have the most news value and best match with two super topics have been selected for labeling. These windows are \{14, 15, 16, 17, 18, 37, 38, 39, 40\}. To perform this action, four experts were used for labeling, and the final label was the result of these people's opinions.

\begin{table}[]
    \centering
    \caption{Statistical information about the datasets}
    \label{tbl:stat datasets}
    \begin{tabular}{|c|c|c|}
        \hline
        name &
        FA CUP &
        Telegram \\
        \hline
        Number of posts &
        189034 &
        10209 \\
        \hline
        Time interval &
        360 &
        32 \\
        \hline
        The size of each window(in minutes) &
        1 &
        720 \\
        \hline
        Number of super topics &
        1 &
        2 \\
        \hline
        Number of topics &
        13 &
        25 \\
        \hline
        Begin &
        14:00:00 05/05/2012 &
        00:00:00 01/01/2017 \\
        \hline
        end &
        20:00:00 05/05/2012 &
        24:00:00 31/01/2017  \\
        \hline
        the whole interval(in hours) &
        6 &
        384 \\
        \hline
    \end{tabular}
\end{table}

\subsection{Parameters Setup}\label{subsec:ParamSetup}
All of the parameters used in this paper are discussed in this section. Table \ref{tbl:ParamSetup} summarizes the parameters used in the framework. This table represents the methods used in this paper, their parameters, their range, and the selected value for each one. $h$ and $\delta$ represent Keyword rating and damping factor, respectively, which are parameters used in HWA methods. Both parameters are tuned in various numbers to find a suitable value. For instance, parameter $h$ has an initial value of 5 and a termination value of 50. This parameter will increase by 5.

\begin{table}[]
    \centering
    \caption{The parameter and their values used in the proposed method. Some of the parameters are tuned to get better results. For example, parameter $h$ from keyword rating is tuned over the range of [5, 50]. To find an optimum value in each run we increase this parameter by 5.}
    \label{tbl:ParamSetup}
    \begin{tabular}{cccc}
        \hline
        Method & Parameters & Span & Value\\
        \hline
        \multirow{2}{*}{HWA} 
            & Keyword Rate(h)   & range( 5, 50, 5)     & 30 \\
            & Damping factor($\delta$ ) & range(0.1, 1.0, 0.1) & 0.5\\
        \hline
        \multirow{1}{*}{Clustering (Hdbscan)} 
            & Min cluster size & & 5\\
        \hline
    \end{tabular}
\end{table}

Table \ref{tbl:hparam} represents experiment results for different values of parameters $h$ and $\delta$. This table illustrates that $h=30$ and $\delta = 0.5$ perform better in keyword-F1 measurement. We examined different values for these parameters for both datasets and all evaluation metrics given in the section \ref{subsec:evaluation_metrics}. Table \ref{tbl:hparam} shows the keyword-F1 measurement for the FA-Cup dataset. finding optimum values for these parameters is the limitation of the proposed method. The optimum value for $h$ in the telegram dataset is 10 however this value has the optimum value of 30 in the FA-Cup dataset as illustrated in the table. To make the method independent of different languages, datasets, and social media platforms, we fixed this value to 30.

In our other paper \cite{khadivi2023GE} we have tested different methods for the clustering part. In that paper, we have used k-means, optics, and HDBSCAN. We find out that HDBSCAN performs better than the other algorithms because it does not need any parameters that affect clustering results. 

\begin{table}[]
    \centering
    \caption{finding the optimum values of parameters $h$ and $\delta$ on keyword-F1 measure on FA-Cup dataset}
    \label{tbl:hparam}
    \begin{tabular}{c|c|c || c c|c|c}
        \hline
        rate & damping factor & F1-measure & & rate & damping factor & F1-measure\\
        \hline
5 &     0.1 & 	0.268791568	& 	& 30 & 	0.1 & 	0.367970667 \\
5 &     0.2 & 	0.268791568	&   & 30 & 	0.2 & 	0.403244385 \\
5 &     0.3 & 	0.267996573 &   & 30 & 	0.3 & 	0.503244385 \\
5 & 	0.4 & 	0.267996573	& 	& 30 & 	0.4 & 	0.5312793 \\
5 & 	0.5 & 	0.267996573	& 	& \textbf{30} & 	\textbf{0.5} & 	\textbf{0.545809} \\
5 & 	0.6 & 	0.267996573	& 	& 30 & 	0.6 & 	0.531022 \\
5 & 	0.7 & 	0.263614197	& 	& 30 & 	0.7 & 	0.5244384 \\
5 & 	0.8 & 	0.263614197	& 	& 30 & 	0.8 & 	0.445657 \\
5 & 	0.9 & 	0.263614197	& 	& 30 & 	0.9 & 	0.432808548 \\
5 & 	1.0 & 	0.263614197	& 	& 30 & 	1.0 & 	0.338547514 \\
10 & 	0.1 & 	0.362656423	& 	& 35 & 	0.1 & 	0.212162431 \\
10 & 	0.2 & 	0.362656423	& 	& 35 & 	0.2 & 	0.214604711 \\
10 & 	0.3 & 	0.272440847	& 	& 35 & 	0.3 & 	0.106705995 \\
10 & 	0.4 & 	0.272440847	& 	& 35 & 	0.4 & 	0.106705995 \\
10 & 	0.5 & 	0.272440847	& 	& 35 & 	0.5 & 	0.105456436 \\
10 & 	0.6 & 	0.273113506	& 	& 35 & 	0.6 & 	0.104144348 \\
10 & 	0.7 & 	0.273113506	& 	& 35 & 	0.7 & 	0.105456436 \\
10 & 	0.8 & 	0.271750483	& 	& 35 & 	0.8 & 	0.105456436 \\
10 & 	0.9 & 	0.271750483	& 	& 35 & 	0.9 & 	0.105456436 \\
10 & 	1.0 & 	0.271750483	& 	& 35 & 	1.0 & 	0.105456436 \\
15 & 	0.1 & 	0.33945484	& 	& 40 & 	0.1 & 	0.15045101 \\
15 & 	0.2 & 	0.33945484	& 	& 40 & 	0.2 & 	0.15045101 \\
15 & 	0.3 & 	0.33945484	& 	& 40 & 	0.3 & 	0.123489172 \\
15 & 	0.4 & 	0.342268278	& 	& 40 & 	0.4 & 	0.123489172 \\
15 & 	0.5 & 	0.33945484	& 	& 40 & 	0.5 & 	0.123489172 \\
15 & 	0.6 & 	0.336536965	& 	& 40 & 	0.6 & 	0.123489172 \\
15 & 	0.7 & 	0.33945484	& 	& 40 & 	0.7 & 	0.123489172 \\
15 & 	0.8 & 	0.33945484	& 	& 40 & 	0.8 & 	0.123489172 \\
15 & 	0.9 & 	0.344984506	& 	& 40 & 	0.9 & 	0.123489172 \\
15 & 	1.0 & 	0.344984506	& 	& 40 & 	1.0 & 	0.123489172 \\
20 & 	0.1 & 	0.15045101	& 	& 45 & 	0.1 & 	0.209600784 \\
20 & 	0.2 & 	0.15045101	& 	& 45 & 	0.2 & 	0.209600784 \\
20 & 	0.3 & 	0.15045101	& 	& 45 & 	0.3 & 	0.104144348 \\
20 & 	0.4 & 	0.15045101	& 	& 45 & 	0.4 & 	0.104144348 \\
20 & 	0.5 & 	0.15045101	& 	& 45 & 	0.5 & 	0.104144348 \\
20 & 	0.6 & 	0.15045101	& 	& 45 & 	0.6 & 	0.104144348 \\
20 & 	0.7 & 	0.154367727	& 	& 45 & 	0.7 & 	0.104144348 \\
20 & 	0.8 & 	0.154367727	& 	& 45 & 	0.8 & 	0.104144348 \\
20 & 	0.9 & 	0.154367727	& 	& 45 & 	0.9 & 	0.104144348 \\
20 & 	1.0 & 	0.15045101	& 	& 45 & 	1.0 & 	0.104144348 \\
25 & 	0.1 & 	0.26450054	& 	& 50 & 	0.1 & 	0.182763064 \\
25 & 	0.2 & 	0.26450054	& 	& 50 & 	0.2 & 	0.182763064 \\
25 & 	0.3 & 	0.274736188	& 	& 50 & 	0.3 & 	0.090287313 \\
25 & 	0.4 & 	0.274736188	& 	& 50 & 	0.4 & 	0.090287313 \\
25 & 	0.5 & 	0.271093443	& 	& 50 & 	0.5 & 	0.090287313 \\
25 & 	0.6 & 	0.363879754	& 	& 50 & 	0.6 & 	0.090287313 \\
25 & 	0.7 & 	0.365951805	& 	& 50 & 	0.7 & 	0.090287313 \\
25 & 	0.8 & 	0.365951805	& 	& 50 & 	0.8 &	0.090287313 \\
25 & 	0.9 & 	0.367970667	& 	& 50 & 	0.9 & 	0.090287313 \\
25 & 	1.0 & 	0.363879754	& 	& 50 & 	1.0 & 	0.090287313 \\
    \end{tabular}
\end{table}

\subsection{Evaluation Metrics}\label{subsec:evaluation_metrics}
To evaluate the correctness of the extracted topics, a metric is needed that can compare the topics in the grand truth (GT) and the system. This paper uses Topic Precision and Topic Recall as performance evaluation scales. F-measure is a combination of Topic Precision and Topic Recall. All three of these criteria have been used to evaluate all windows.
\begin{itemize}
    \item \textbf{Topic Precision}\\
    The number of extracted Topics that match the Topics in GT.
    \begin{equation}
        \label{eq: TopicPrecision}
       Topic\ Precision=\frac{\mid topics\ matches\ to\ GT\mid}{\mid extracted\ topics\mid}
    \end{equation}
    
    \item \textbf{Topic Recall}\\
    The number of correctly extracted Topics from GT Topics.
    \begin{equation}
        \label{eq: TopicRecall}
       Topic\ Recall=\frac{\mid successfully\ detected\ GT\ topics\mid}{\mid GT\ topics\mid}
    \end{equation}
    
    \item \textbf{Topic F1-Measure}\\
    Using the above concepts, F-measure is defined as follows
     \begin{equation}
        \label{eq: TopicF1}
       F_1 measure=2\times\frac{Topic\ Precision\times Topi c\ Recall}{Topic\ Precision+Topic\ Recall}
    \end{equation}
\end{itemize}

For example, it is assumed that four GT topics are presented if 6 out of 10 extracted topics are matched with 3 GT topics, then $Topic Precision=3/4$ and $Topic Recall=6/10$.

In the FA-Cup dataset, the works done so far were limited to reporting the Top-k Topic recall rate. So in this paper, only the same criterion is reported for the FA-Cup dataset. On the other hand, all three metrics have been reported for the Telegram dataset.
Another metric, reported by previous studies for the FA-Cup dataset, is based on keywords (Top-2 keyword precision, Top-2 keyword recall, and Top-2 keyword-F1 measure), also applied to the proposed method. The results obtained for the proposed method using the introduced evaluation criteria are given below.


\subsection{Results}\label{subsec:results}
The efficiency of the proposed solution has been evaluated on both FA-Cup and Telegram datasets. The results obtained from other methods have been used to evaluate and compare the proposed method with others. The results of various top-K tests on the proposed method also have been added. The results of this evaluation are described below.

\subsubsection{FA-Cup dataset}\label{subsec:results_facup}
In this section, the proposed topic detection algorithm has been implemented on each time window for at least one topic in the ground truth. The performance results of the proposed solution with the compared methods are given on the FA-Cup dataset. The announced results are based on the values announced in the compared papers. Table \ref{tbl:results_TopK_topicsrecall} shows the results based on top-k topic-recall. As seen in the table, in topic-recall, the best-reported method found all topics in the Top 8, while in the proposed method, this value is Top 6. Also, this value has increased by 0.015 and 0.036. In the discussion of keyword F-1 measure, as the comparative results in Table \ref{tbl:results_F1_measure} show, few papers have reported this measure in their works; however, compared to the reported works, the proposed method has a significant improvement, which is since in keyword-recall, the proposed method was able to recover all the keywords. It should be noted that because some methods did not report their keyword-recall rate, these studies are not included in the comparison table.

\subsubsection{Telegram dataset}\label{subsec:results_Telegram}
\begin{figure}
    \centering
    \includegraphics[width=0.9\textwidth]{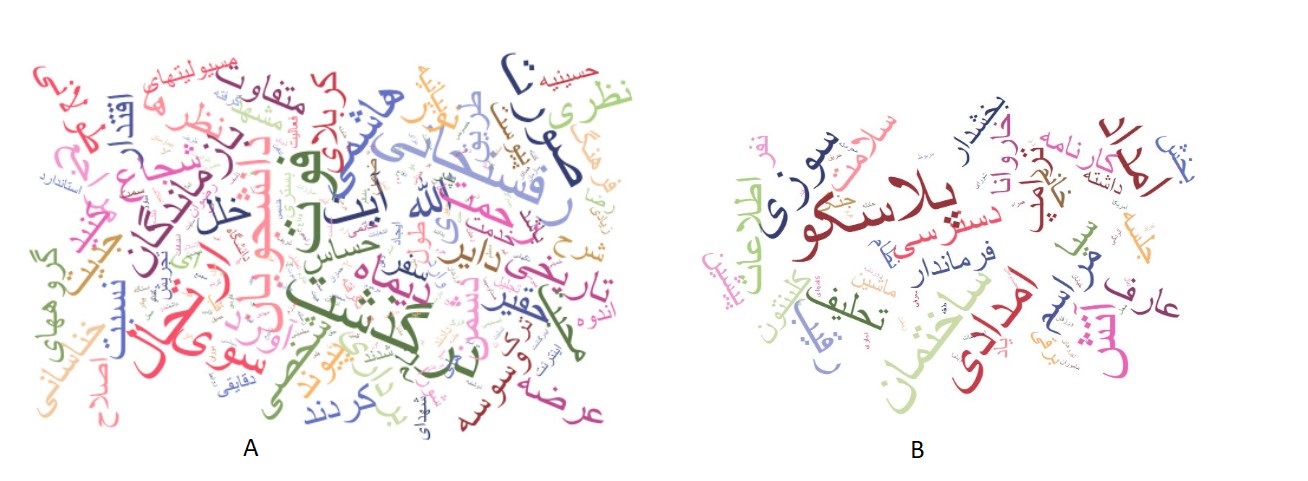}
    \caption{The word tag of topics in two windows 16 and 37 (A and B). Window 16 includes the super topic "Death of Ayatollah Hashemi" and window 37 includes the super topic "Plasko building fire incident". In addition, each window contains other topics. For example, in window 37, the topic "Inauguration of the President of the United States" can also be seen }
    \label{fig:WordTag}
\end{figure}

The methods' results for comparing the topic detection performance based on F-measure are given in Table \ref{tbl:results_F1_measure_telegram} to evaluate the proposed model's performance. To enable an equal comparison, the authors have implemented all the methods given in the table. These algorithms are applied to the dataset introduced in section \ref{subsec:Dataset_Telegram}, and their results are compared with GT using the evaluation criteria introduced. This table shows that the proposed HWA-based method performs best in the F-measure's mean value. Also, considering that two windows 16 and 37 are the windows where the two super hot topics, "the death of Ayatollah Hashemi Rafsanjani" and "the fire of the Plasco building", occurred for the first time, they are more critical as a result. These types of windows, called interference windows, have many topics that have already occurred and are being published, as well as those that have just occurred. For this, most algorithms may get confused in these types of windows. For example, as can be seen in the table, the HUPM algorithm is confused. However, the proposed algorithm received the highest score in these windows compared to other algorithms, which shows the high accuracy of this algorithm. Figure \ref{fig:WordTag} shows the super words of each of these windows based on the extracted topics. Window 16 includes the super topic "Death of Ayatollah Hashemi", and window 37 includes the "Plasko building fire incident". In window 37, the keywords \{fire, building, Plasco\} (
\{ \setcode{utf8} \<پلاسکو>, \<ساختمان>, \<آتش> \}
) indicate their topics, which are displayed more prominently in the cloud due to the frequency of their occurrence. Also, this window includes two topics, "Inauguration of Donald Trump" and "Kharwana governor's election" which are respectively shown with the keywords \{Trump, inauguration, rival, Clinton\} (
\{ \setcode{utf8} \<رقیب>, \<تحلیف>, \<ترامپ> \ , \<کلینتون>\}
) and \{mayor, Kharwana, governor\} (
\{ \setcode{utf8} \<بخشدار>, \<خاروانا>, \<فرماندار> \}
)in the super words.

 \begin{table}[]
     \centering
     \caption{Top-K topics-recall comparison in FA-CUP dataset}
     \label{tbl:results_TopK_topicsrecall}
     \begin{tabular}{p{1.8cm}cccccccccc}
         \hline
         \multirow{2}{*}{\textbf{Method}}  &\multicolumn{10}{c}{\textbf{TOP-K topic recall}} \\
             &2 &4 &6 &8 &10 &12 &14 &16 &18 &20  \\
         \hline
         LDA\cite{Blei2003} &0.692 &0.692 &0.84 &0.84 &0.92 &0.92 &0.84 &0.84 & 0.84 &0.75\\
         Doc-P\cite{FSD} & 0.769 & 0.85  & 0.92  & 0.92  &1   &1   &1  &1   &1  &1 \\
        Gfeat-P\cite{tweetmotif} & 0     & 0.308 & 0.308 & 0.375 & 0.375 & 0.375 & 0.375 & 0.375 & 0.375 &0.375 \\
        SFPM\cite{SFPM}  &0.615 & 0.84  & 0.84  & 1     & 1     & 1     & 1     & 1     &1  &1  \\
        BNGram\cite{Aiello2013} &0.769 & 0.92  & 0.92  & 0.92  & 0.92  & 0.92  & 0.92  &0.92  &0.92  &0.92  \\
        SVD+Kmean\cite{SVDKmeans} &0.482 & 0.596 & 0.71  & 0.824 & 0.938 & 0.951 & 0.951 & 0.951 &0.951 & 0.951 \\
        SNMF-Orig\cite{SNMF} & 0.1   & 0.177 & 0.254 & 0.331 & 0.389 & 0.389 & 0.389 & 0.389 &0.389 & 0.389 \\
        SNMF-KL\cite{SNMF} & 0.167 & 0.334 & 0.502 & 0.67  & 0.837 & 0.837 & 0.84  & 0.85  &0.85  &0.924 \\
        Exemplar\cite{Exemplar-Based} & 0.81  & 0.838 & 0.886 & 0.908 & 0.916 & 0.916 & 0.916 & 0.916 &0.916 &0.916 \\
        EHG\cite{Saeed2019} & 0.379 & 0.591 & 0.727 & 0.727 & 0.864 & 0.864 &1  &1     &1     &1  \\
        TopicBERT\cite{TopicBert} & 0.81  & 0.951 & 0.951   & 1  & 1     & 1     & 1     & 1     & 1  & 1   \\
        SMM \cite{SMM} & 0.538  & 0.923 & 0.923   & 0.923 & 1     & 1     & 1     &1     &1  &1   \\
        \textbf{HWA based (proposed)} &\textbf{0.846} &\textbf{0.966} &\textbf{1}     &\textbf{1}     &\textbf{1}     &\textbf{1}     &\textbf{1}     &\textbf{1}     &\textbf{1}     &\textbf{1}   \\
         \hline
     \end{tabular}
\end{table}

\begin{table}[]
    \centering
    \caption{Top-2 keyword-precision, Top-2 keyword-recall and Top-2 keyword-F1 measure comparison in FA-Cup dataset}
    \label{tbl:results_F1_measure}
    \begin{tabular}{lccc}
        Method               &keyword-precision & keyword-recall & F-1 measure \\
        \hline
        LDA \cite{Blei2003}       & 0.164             & 0.683          & 0.264       \\
        Doc-P \cite{FSD}          & 0.337             & 0.583          & 0.427       \\
        Gfeat-P \cite{tweetmotif} & 0.000             & 0.000          & 0.000       \\
        SFPM \cite{SFPM}          & 0.234             & 0.658          & 0.345       \\
        BNGram \cite{Aiello2013}  & 0.299             & 0.578          & 0.394       \\
        HUPM \cite{HUPM}          & 0.370             & 0.606          & 0.459       \\
        \textbf{HWA based (proposed)} & \textbf{0.375}            &\textbf{1}              &\textbf{0.545}  \\
        \hline
    \end{tabular}
\end{table}

\begin{table}[]
    \centering
    \caption{Topic precision, Topic recall and Topic F1 measure comparison in Telegram dataset}
    \label{tbl:results_F1_measure_telegram}
    \begin{tabular}{lccc}
        Method               &Topic Precision & Topic Recall & F-1 measure \\
        \hline
        Twevent \cite{twevent}       & 0.527273 & 0.413793 & 0.463691 \\
        HUPC \cite{HUPC}                      & 0.908046 & 0.293103 & 0.443161 \\
        HUPM \cite{HUPM}                      & 0.941176 & 0.172414 & 0.291439 \\
        AGF(base) \cite{AGF_base}             & 0.516484 & 0.827586 & 0.63603 \\
        \textbf{HWA based(proposed)} & 0.838889 & 0.561965 & \textbf{0.673055} \\
        \hline
    \end{tabular}
\end{table}

\section{Conclusion}\label{sec:conclusion}
This paper proposes a topic detection framework using the Human Word Association (HWA) method combined with a word embedding model and a clustering method. HWA imitates human performance in understanding the relationships between words. This framework is applied to social media posts and evaluated using two datasets. The first dataset is the FA-Cup dataset, a benchmark in English for topic detection. The second dataset comprises posts collected from Telegram in Persian.

After preprocessing, keywords and their co-occurrences are extracted from the posts. HWA values are then calculated, and AGF (Association Graph Frequency) values are extracted. Using AGF, initial patterns are created. To avoid generating similar patterns, word embedding is employed to create a vector for each pattern. These vectors are then clustered to extract topics.

The results on the FA-Cup dataset were evaluated based on the Topic-recall criterion, and the proposed method successfully extracted all topics within the Top-6 range, compared to the previous best method, which did so in the Top-8 range. Additionally, the proposed method showed an increase in Topic-recall at lower GAs compared to the previous method. In terms of the keyword F1 measure, the proposed method improved by 0.089.

For evaluating the method on the Telegram dataset, previous methods were implemented by the author of this paper due to the change in the dataset and language. The results were also evaluated based on the Topic F1 measure. The main reason for using this dataset was to investigate the impact of different media and languages on the proposed method. While the FA-Cup dataset serves as a benchmark, keyword rating, co-occurrence, and AGF modules operate independently of language. However, the clustering module depends on the language model and its embeddings. By using a strong and suitable model, this issue is addressed.

In the proposed model, the fastText model was used because it supports multiple languages, including Farsi and English, and effectively handles out-of-vocabulary (OOV) words. The obtained results demonstrate that the proposed method is independent of language. Consequently, this method can be expanded to all text-based platforms, allowing for topic detection across various media.

The proposed framework has outperformed other methods in this approach. For future work, HWA could be utilized in the form of a graph, with the resulting data analyzed through graph processing. Another suggestion is to integrate the latest large language models, such as BERT and GPT, into the framework.


\section*{Declarations}
\subsection*{Funding}
No funds, grants, or other support were received.
\subsection*{Conflict of interest}
The authors have no conflicts of interest to declare that are relevant to the content of this article.
\subsection*{Availability of data and materials}
The dataset generated during the current study, Sep\_TD\_Tel01, is available in the Mendeley repository, \url{https://doi.org/10.17632/372rnwf9pc}.
\subsection*{Ethics approval}
Not applicable.
\subsection*{Authors' contributions}


\bibliography{sn-bibliography}


\end{document}